\documentclass{sig-alternate-05-2015}
\usepackage{color}
\usepackage{ifpdf}

\usepackage{perpage} % Kin addition: Makes figure appear in order
\MakeSorted{figure} % Kin addition: Makes figure appear in order

\begin{document}

% Copyright
%\setcopyright{acmcopyright} % commented out by Dotun

\title{Early Detection of Combustion Instabilities using Deep Convolutional Selective Autoencoders on Hi-speed Flame Video
%\titlenote{(Produces the permission block, and
%copyright information). For use with
%SIG-ALTERNATE.CLS. Supported by ACM.}
}
%\subtitle{[Extended Abstract]
%\titlenote{A full version of this paper is available as
%\textit{Author's Guide to Preparing ACM SIG Proceedings Using
%\LaTeX$2_\epsilon$\ and BibTeX} at

\numberofauthors{4} %  in this sample file, there are a *total*
\author{
% 1st. author
\alignauthor
Adedotun Akintayo\\
       \affaddr{Mechanical Engineering Department}\\
       \affaddr{Iowa State University}\\
       \affaddr{Ames, IA 50011}\\
       \email{akintayo@iastate.edu}
% 2nd. author
\alignauthor
Kin Gwn Lore\\
       \affaddr{Mechanical Engineering Department}\\
       \affaddr{Iowa State University}\\
       \affaddr{Ames, IA 50011}\\
       \email{kglore@iastate.edu}
% 3rd. author
\alignauthor
Soumalya Sarkar\\
       \affaddr{Decision Support and Machine Intelligence}\\
       \affaddr{United Technologies Research Center}\\
       \affaddr{East Hartford, CT 06118}\\
       \email{sms388@gmail.com}
\and  % use '\and' if you need 'another row' of author names
% 4th. author
\alignauthor Soumik Sarkar\\%\titlenote{corresponding author}\\
       \affaddr{Mechanical Engineering Department}\\
       \affaddr{Iowa State University}\\
       \affaddr{Ames, IA 50011}\\
       \email{soumiks@iastate.edu}
}
% There's nothing stopping you putting the seventh, eighth, etc.
% author on the opening page (as the 'third row') but we ask,
% for aesthetic reasons that you place these 'additional authors'
% in the \additional authors block, viz.
%\additionalauthors{Additional authors: John Smith (The Th{\o}rv{\"a}ld Group,
%email: {\texttt{jsmith@affiliation.org}}) and Julius P.~Kumquat
%(The Kumquat Consortium, email: {\texttt{jpkumquat@consortium.net}}).}
% Just remember to make sure that the TOTAL number of authors
% is the number that will appear on the first page PLUS the
% number that will appear in the \additionalauthors section.

\maketitle
\begin{abstract}
This paper proposes an end-to-end convolutional selective autoencoder approach for early detection of combustion instabilities using rapidly arriving flame image frames. The instabilities arising in combustion processes cause significant deterioration and safety issues in various human-engineered systems such as land and air based gas turbine engines. These properties are described as self-sustaining, large amplitude pressure oscillations and show varying spatial scales periodic coherent vortex structure shedding. However, such instability is extremely difficult to detect before a combustion process becomes completely unstable due to its sudden (bifurcation-type) nature. In this context, an autoencoder is trained to selectively mask stable flame and allow unstable flame image frames. In that process, the model learns to identify and extract rich descriptive and explanatory flame shape features. With such a training scheme, the selective autoencoder is shown to be able to detect subtle instability features as a combustion process makes transition from stable to unstable region. As a consequence, the deep learning tool-chain can perform as an early detection framework for combustion instabilities that will have a transformative impact on the safety and performance of modern engines.

\end{abstract}


\section{Introduction}

\begin{figure}[ht]
\vskip 0.1in
\begin{center}
\centerline{\includegraphics[width=\columnwidth]{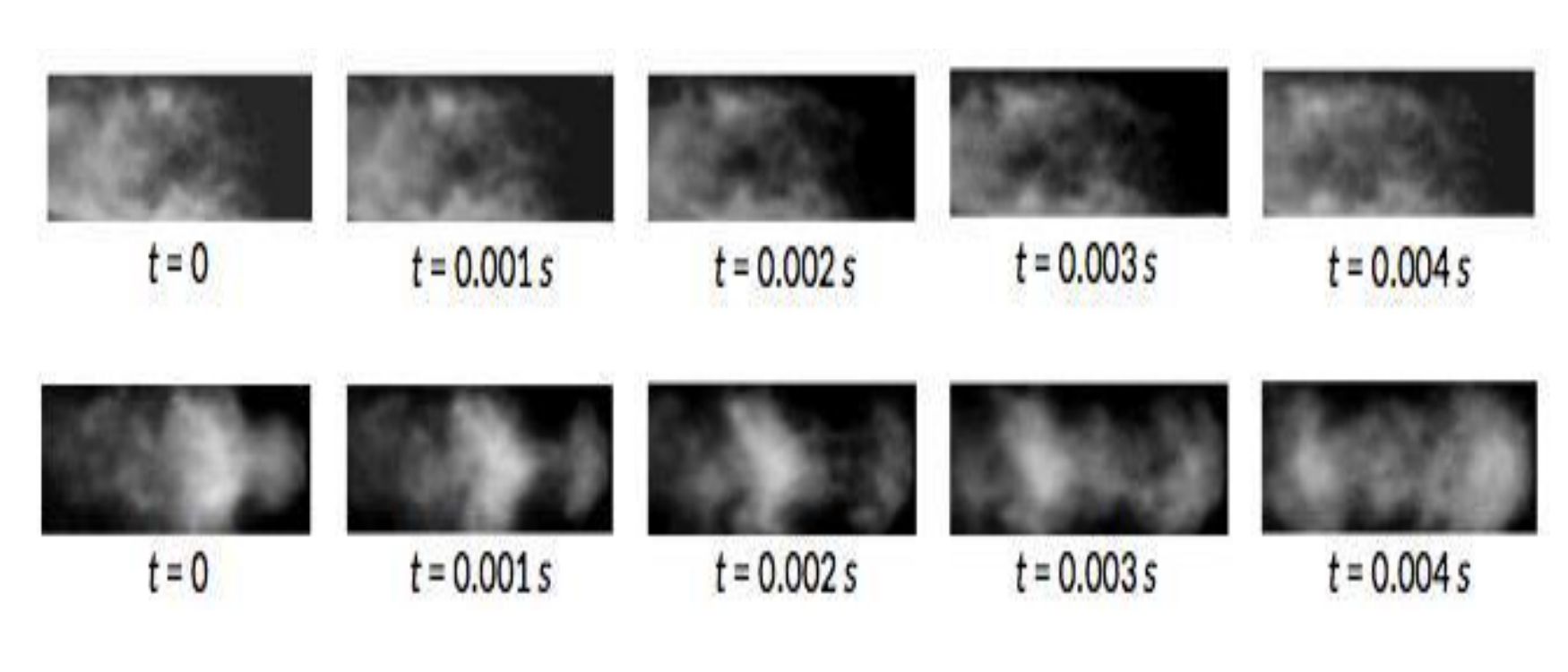}}
\caption{Greyscale images of gradual time-varying development of instability structure at two different parameter values}
\label{fig:example_dataset}
\end{center}
\vskip -0.1in
\end{figure}

Deep Learning models have been shown to outperform all other state-of-the-art machine learning techniques for handling very large dimensional data spaces and learn hierarchical features in order to perform various machine learning tasks. However, most of the studied applications primarily have been in the domains of image, speech and texts processing. For example, convolutional neural network-based applications include Graph Transformer Networks (GTN) for rapid, online recognition of handwriting~\cite{LBBH98}, natural language processing~\cite{CW08}, large vocabulary continuous speech recognition~\cite{SPKL15} and avatar CAPTCHA machine image recognition~\cite{BC12} by training machines to distinguish between human faces and computer generated faces. In this paper, we propose a novel selective autoencoder approach within a deep convolutional architecture for implicit labeling in order to derive soft labels from extreme classes that are explicitly labeled as either positive or negative examples. This particular property is significant for tracking continuous temporal phenomenon such as the transition from combustion stability to instability, where labels of extreme states (stable or unstable) are available but intermediate state labels are not. Explicit labels are utilized to selectively mask selective features while allowing other features to remain. Figure~\ref{fig:example_dataset} shows greyscale images describing typical gradual development of instability at the stated parameters in the swirl-stabilized combustor used for the experiment.

Labeling (e.g., structured and implicit) can be considered a multi-class classification problem ~\cite{HE10}. For example, three stage Hidden Markov Models (HMM) were used for handling speech recognition~\cite{R89} problems, parts of speech tagging~\cite{AM12} and sequence labeling because they derive the relationships from observations to state and state to state in dynamic systems. Maximum Entropy Markov Model (MEMM) a discriminative modification of HMM was introduced to overcome the latter's recall and precision problems especially in labeling texts. In those models, conditional probability of the desired labels are learnt directly based on the uncertainty maximization idea. Applications of MEMM for natural language processing can be found in~\cite{BPP96}.

Due to "label bias" defects of MEMM, a Conditional Random Field (CRF), which is a joint Markov Random Field (MRF) of the states conditioned on the whole observations was later explored by ~\cite{LMP01}. It enabled considering the global labels of the observation as against localization of labels of MEMM~\cite{HE10}. However, labeling in this case is made computationally complex by the relaxation of statistical independence assumption of the observations which most of the models assume.

Recurrent Neural Networks (RNNs) have been utilized for sequence labeling problems due to its cyclic connections of neurons~\cite{AG14} as well as its temporal modeling ability. Although earlier construction of RNNs was known to have short ranged memory issues and a restrictive unidirectional information context access, formulation of a bidirectional Long Short Term Memory (LSTM)~\cite{AGJS05} resolved such issues. However, this construction adds to the complexity of the model significantly as typically two RNNs get connected through same output layer.

From the application standpoint, early detection of instability in the combustion chambers of dynamic systems aids anticipative actions for reducing its consequent effects. Visualizing the features characterizing the intermediate frames of its spectrum is an important approach to unravel the processes that precedes instability. The authors ~\cite{SLSRCPR15} introduced Deep Belief Networks, DBNs as a viable technique to achieve the aim with a view to exploring other machine learning for confirmation. They improved on that by applying a modular neural-symbolic approach ~\cite{NIPS15} in another publication. 

%In this paper, an end-to-end selective autoencoder is developed to not only visualize the features, but also the results from frames in test sets. Furthermore, since the instability patterns are usually some discrete bursts, pseudo-results are used to describe the interpolation technique for implicit labeling. Based on that, attack of a conceptualized neighborhood adjacency problem may then be introduced via a graduated linear path to reduce the effects of the curse of dimensionality. 

In this paper, we propose a deep convolutional selective autoencoder-based anomaly detection framework for the crucial physical process of combustion where a pure black box model is unacceptable in order to enable domain interpretation and better understanding of the underlying complex physics. Combustion instability is a significant anomaly characterized by high-amplitude flame oscillations at discrete frequencies that reduces the efficiency and longevity of aircraft gas-turbine engines. Full-blown instability can be differentiated from stable combustion via video analysis with high confidence because unstable combustion flames show distinct coherent structures similar to 'mushroom' shapes. But it is extremely difficult to detect an onset of instability early due fast spatio-temporal transience in the video data. Therefore, the instability detection problem boils down to an implicit soft labeling problem where we train a deep model using hi-speed flame videos with explicit labels of stable and unstable flames such that it recognizes onset of instability early as the combustion process makes transition from a stable to unstable region.

%are faced with how to determine the proximity in states of two adjacent observations for generic applications from already examined spatio-temporal observations perspective.
Conceptually, this is similar to cognitive psychologists' description of human reasoning in object classification~\cite{TKGG11}. An example is to consider how a child is taught on intrinsic classes. A similar problem is detecting a cross breed of dog and wolf including how close the animal is to either of the classes.

\noindent\textbf{Contributions:} The main contributions of this paper is delineated below:
\begin{itemize}
\item a convolutional selective autoencoder framework based on emerging deep learning techniques is proposed for early detection of combustion instability,
\item the method avoids extensive expert-guided feature handcrafting~\cite{FCNL13} while addressing a complex physical phenomenon like combustion to discover coherent structures in the images,
%\item be intrinsically bidirectional~\cite{AGJS05},
\item the proposed framework is able to learn from high dimensional datasets (e.g, high speed video) of most applications and provide a platform for determining the degree of relationship between the states of two temporally close observations,
\item A metric to desired level of granularity is constructed to track the onset of combustion instability and detect pre-transition phenomena such as 'intermittence'. Intermittence is a temporary (of the order of millisecond) blast of instability characterized by small and partially observable coherent structure.
\item extensive validation is provided based on laboratory-scale combustion data collected under various realistic operating conditions.
%\item be able to visualize the outcome in tandem with the values of the chosen metric for the labels
\end{itemize}

%\subsection{Paper layout}
%\label{sub:layout}
\noindent\textbf{Paper organization:} The paper was introduced earlier in this section with an idea of the motivating problem. The next section is set apart to discuss: prior work on the proposed approach, the chosen problem, and an essential part of the attack technique. In section ~\ref{sec:method}, the main architecture for the problem is discussed followed by a measure of similarity to access the result quantitatively. Section ~\ref{sec:implementation} provided an opportunity to introduce the problem dataset collection and then the implementation of the composite architecture. The results obtained for the hypothesis is discussed in section~\ref{sec:results} after which we conclude the paper in section~\ref{sec:conclude} as well as give some insights to the direction of future works.

\section{Background}\label{sec:priorwork}
This section provides a brief overview of convolutional networks, a description of the example problem of detecting combustion instability, and the notion of implicit labeling.

\subsection{Convolutional networks}
Convolutional networks~\cite{KSH12} are a type of deep networks that offer discriminative advantages as in the MEMM as well as providing globally relationship between observations as in the CRF. The architectures rely primarily on local neighborhood matching for data dimension reduction using nonlinear mapping (i.e. sigmoid, softmax, hyperbolic tangent). Each unit of the feature maps has common shared weights or kernels for efficient training in relatively - compared to fully connected layers - lower trainable parameters, added to an additive bias on which is squashed. Feature extraction and classifier learning are the two main functions of these networks~\cite{LBBH98}. However, to learn the most expressive features, we have to determine the invariance rich codes embedded in the raw data and then follow with a fully connected layer to reduce further the dimensionality of the data and map the most important codes to a low dimension of the examples. Many image processing and complex simulations depend on the invariance property of the convolution neural network stated in~\cite{LB95} to prevent overfitting by learning expressive codes.

%since units at different receptive fields are usually invariance of other similar but shifted, scaled and distorted units at other receptive fields.
The feature maps are able to preserve local neighborhood patterns for each receptive field as with over-complete dictionaries in~\cite{AEB06}. The fully connected layers tend to complement the learned features by propagating only the highly active weights and serving as the classifier. A more detailed review may be found in~\cite{LBBH98} where the authors note the advantage of local correlation enforcing convolution before spatio-temporal recognition. For efficient learning purposes, convolutional networks are able to utilize distributed map-reduce frameworks~\cite{FM04} as well as GPU computing.

\subsection{The problem of combustion instability}
Combustion instability reduces the efficiency and longevity of aircraft gas-turbine engines. It is considered a significant anomaly characterized by high-amplitude flame oscillations at discrete frequencies. These frequencies typically represent the natural acoustic modes of the combustor. Combustion instability arises from a positive coupling between the heat release rate oscillations and the pressure oscillations. Coherent structures are fluid mechanical structures associated with coherent phase of vorticity~\cite{H83}. These structures, whose generation mechanisms vary system wise, cause large scale velocity oscillations and overall flame shape oscillations by curling and stretching. These structures can be caused to shed/generated at the duct acoustic modes when the forcing (pressure) amplitudes are high. There is a lot of recent research interest on detection and correlation of these coherent structures to heat release rate and unsteady pressure. The popular methods resorted for detection of coherent structures are proper orthogonal decomposition (POD)~\cite{BHL93} (similar to principal component analysis~\cite{B06}) and dynamic mode decomposition (DMD)~\cite{S10}, which use tools from spectral theory to derive spatial coherent structure modes.

\subsection{Implicit labeling}
Semi-supervised training for classification takes advantage of the labels at the final layers. A variant of structured labeling by~\cite{KACFC14} called implicit labeling is used to derive soft labels from extreme classes that are explicitly labeled as either positive or negative examples. Explicit labels usually can be utilized to selectively mask one feature, especially that one is not interested in while leaving parsing the class of interest. However, explicit labels on its own can only serve as a classifier for intrinsic classes in the test sets learnt from the training set.

\begin{figure}[ht]
\vskip 0.1in
\begin{center}
\centerline{\includegraphics[width=\columnwidth]{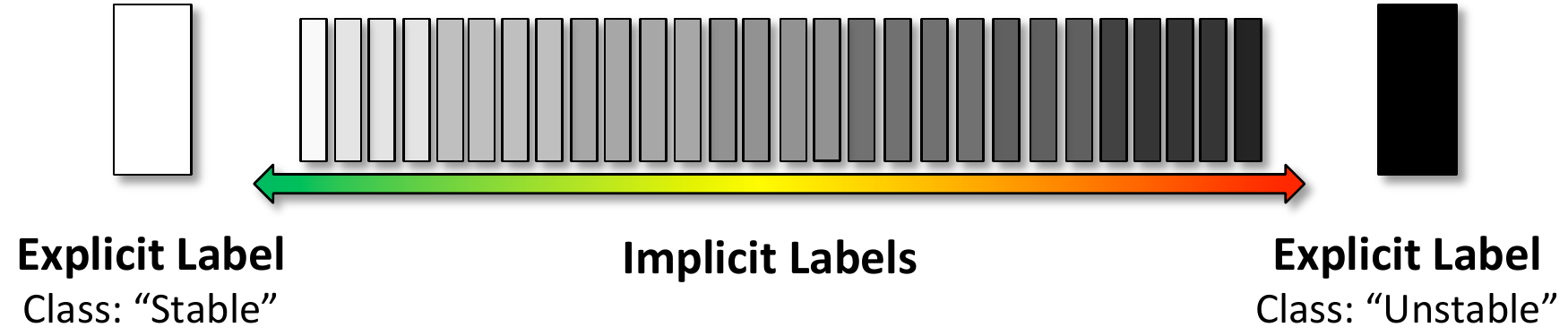}}
\caption{Illustration of implicit method of generating soft labels}
\label{fig:implicit}
\end{center}
\vskip -0.1in
\end{figure}

Implicit labeling here also bears similarity to the sequence labeling~\cite{HE10} with an extra constraint of utilizing prior knowledge provided only by explicit label. It is then fused with convolutional auto-encoder architecture algorithm described in section~\ref{sec:method} to determine the intermediate or transition phases--a mixed breed of a dog and a wolf for instance--and more importantly to what degree is the animal is dog or a wolf. Thus, it attempts to derive soft labels from expert-informed, hard-mined labels as illustrated in fig.~\ref{fig:implicit} with a composite architecture.

\section{Convolutional Selective \\Autoencoder}
\label{sec:method}
\begin{figure*}[!htb]
\centering
 \includegraphics[width=1.0\textwidth,trim={0 0 10 0}]{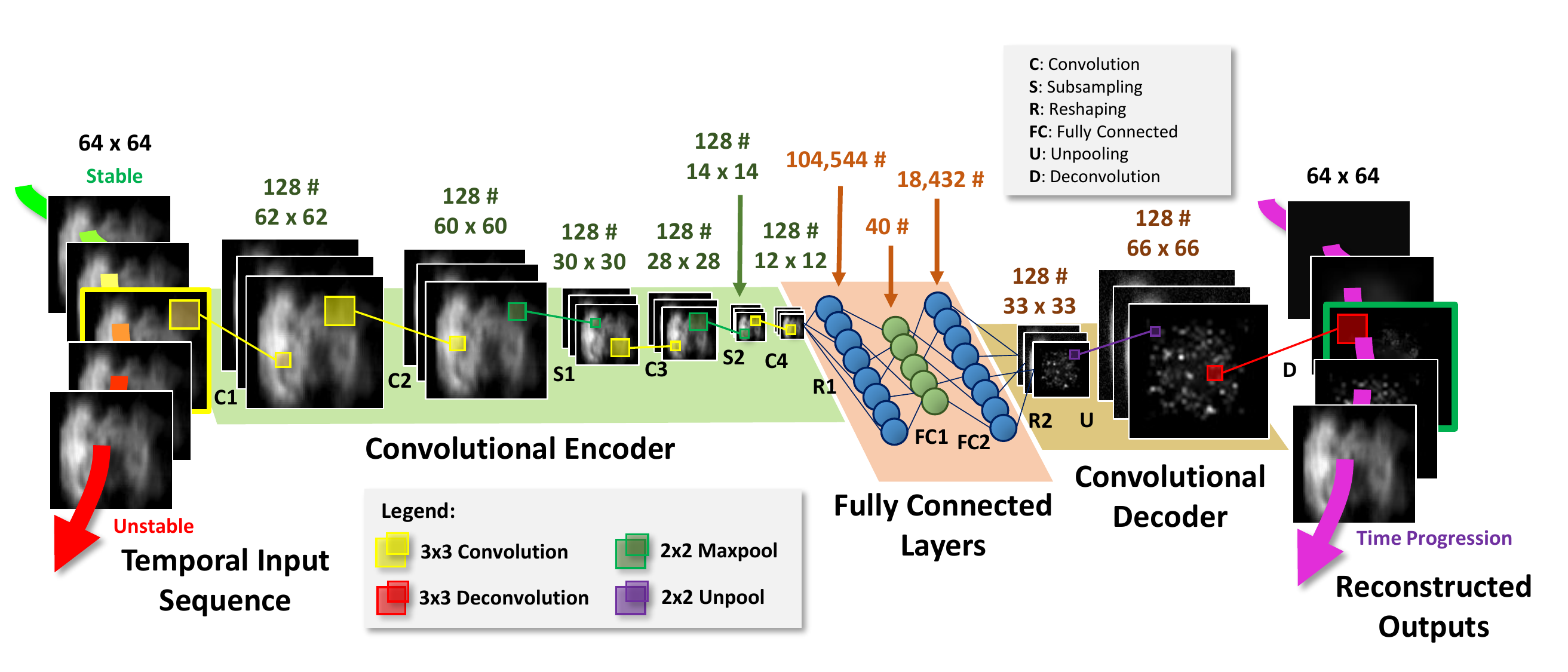}
\caption{Structure of the convolutional autoencoder with associated layer parameters. Best viewed on screen, in color. The encoder portion extracts meaningful features from convolution and sub-sampling operations, while the decoder portion reconstructs the output into the original dimensions through deconvolution and upsampling.}
\label{fig:cae}
\end{figure*}

Based on the convolutional network's (\textit{convnet} for short) performances on several similar tasks reviewed, it was found a suitable candidate for the composite architecture to examine our hypothesis of soft label generation. Having previously utilized convnet architecture, an end-to-end convolutional auto-encoder (as shown in fig.~\ref{fig:cae}), designed and tested to examine another perspective to the current problem instead of adding a symbolic graphical model such as STSA~\cite{NIPS15} at the top level.
The constituent steps for the model to learning from the data are outlined below.\\

\noindent\textbf{Explicit labels and pre-processing}: Given an $M \times N$ dimensional image frames and corresponding ground truth labels (one of the two classes), explicit labels are generated by selectively masking frames with the undesired class with black pixels. Hence, $N$ pairs of input-output pairs $\{(X_i, Y_i)\}$ for $i=1,2,...,N$ are generated where $X$ represents the original images, $Y$ are the masked frames that are considered explicitly as ground truth. The images are then normalized where pixel intensities have zero mean and a standard deviation of 1 as preprocessing.\\

\noindent\textbf{Convolutional layers}: Convolutional autoencoders start with propagation from the input layer to the convolution layer while the penultimate step to the output layer is the deconvolution layer. At each convolution or deconvolution layer, a chosen $(c \times c)$ filter size is convolved with the patches to learn a $z_o-$dimensional feature map from which joint weight over the $z_i-$dimensional feature maps that are useful for enforcing local correlation is learnt to characterize all maps as follows:
\begin{equation}
\hat{Y}_{z_o(m-c+1)(n-c+1)} = C[X_{z_i m n} \star W_{z_i c c} + b_c ]
\end{equation}
where $C$ is the squashing function, $\star$ denoting the convolution operator of the joint weights, $W_{z_i c c} $, $b_c$ the biases and input from previous layer $X_{z_i m n}$. To enhance the invariance further, pooling is done for representative features selection in a local neighborhood. While other pooling schemes exist~\cite{SMB10}, we chose the highly activated units to take up the features for the local  $p \times p$ neighborhood in the maxpooling case. It is expressed as:
\begin{equation}
\hat{Y}_{z_i k l} = \underset{\substack{i \in I; \ i \rightarrow i+p \\ j \in J; \ j \rightarrow j + p }} {\max} (\hat{Y}_{z_i ij})
\end{equation}
where $z_i$ is the number of receptive fields of the input feature maps, $\hat{Y}_{z_i k l}$ is the pooled feature map, $\hat{Y}_{z_i ij}$ is the input from a previous layer, and $I = \{1+(k-1)(p+1), \cdots \}$,  $J = \{1+(l-1)(p+1), \cdots \}$ where $i = 1,2,...,h$ and $j=1,2,...,v$ and $h,v$ denote the horizontal and vertical input dimensions respectively.\\
% $I = \{1, (1+(k-1)p)+1, \cdots \}$ and $J = \{1, (1+(l-1)p)+1, \cdots \}$. $k$ and $l$ are each ceil functions of the p divisor of the respective last two dimensions of the input.\\

\noindent\textbf{Fully connected layers}: Feature maps from previous layers are flattened into row vectors and are passed through the bottleneck layer. The encoding layer encodes the most important feature from the input of the previous layer with the following expression:
\begin{equation}
\hat{Y}_e= E[W_e \hat{Y}+ b_e ]
\end{equation}
and the decoder is given by:
\begin{equation}
\hat{Y}_d= D[W_d \hat{Y}_e+ b_d]
\end{equation}
where $E$ and $D$ stands for the encoder and decoder functions respectively, which are both the same nonlinear function known as the rectified linear unit (ReLU). $b$ denotes the biases and $W$ denotes the weights of the layer. The subscripts $e$ and $d$ indicates the encoder and decoder.
% In addition, to avoid identity learning at this stage, the input to the layer is corrupted with a random Gaussian noise ~\cite{VLB08} and for robustness to distortions in the layer's inputs.\\
The ReLU nonlinearity performed on the argument $f$ is given by:
\begin{equation}
\text{ReLU}(f) = \text{max}(0,f)
\end{equation}
Intuitively, it has the advantage of easier training compared to other nonlinearity types because the activations of each neuron is a piece-wise linear function of argument $f$ and do not saturate.\\ %maximizing the likelihoods whenever there is an \textbf{egg patch}.\\ %Change this dotun

\noindent\textbf{Unpooling}: In this layer, a reversal of the pooled dimension is done by stretching and widening~\cite{SJ15} the identified features from the filters of the previous layer. It is also an upscaling of the feature maps around the axes of symmetry where the reconstructed feature maps are optimized through the back-propagation algorithm.\\

\noindent\textbf{Error minimization}:
This phase is akin to a feedback stage in a control paradigm or a scenario where a teacher--labeled data--provides feedback in performance measure on how well a student--the machine--has learned features related to a particular application--the task. The process included a regularization function as in~\cite{LBBH98} to avoid overfitting the data. The Nesterov momentum-based~\cite{SMDH13} stochastic gradient descent was used for improved results when compared to other loss functions such as adaptive subgradient (ADAGRAD)~\cite{DHS11} and adaptive learning rate method (ADADELTA)~\cite{MZ12} for the reconstruction error updates given the reconstructed output $\hat{Y}_{mn}$ and the labels, ${Y}_{mn}$. Let $\theta = \{\textbf{W}, \textbf{b}\}$ be the set of weights and biases for all layers that are to be optimized by minimizing the loss function $L(\theta)$. The loss function is expressed as:
\begin{equation}\label{equ:tot_loss}
L(\theta) = L_{train}(\theta) + \sigma R(W)
\end{equation}
where $\sigma$ is a parameter controlling the regularization function $R(W)$. $R(W)$ that has the following form:
\begin{equation}
R(W) = (\sum_l \sum_{W_{dim}} W_{l}^2)^\frac{1}{2} +\sum_l \sum_{W_{dim}} |W_{l}|
\end{equation}
where $l$ represents the layer and $W_{dim}$ represents the dimension of the weights at each layer. The works in~\cite{YB08} points out that that SGD with early stopping is equivalent to an $\ell_2-$ regularization. The mean square error training loss is given as:
\begin{equation}
L_{train}(\theta) = \frac{1}{mn}\sum_{i=1}^m \sum_{j=1}^n (Y_{ij} - \hat{Y}_{ij})^2
\end{equation}
Subsequently, the weights are updated for each time step via stochastic gradient descent~\cite{LBBH98}:
\begin{equation}
W_t = W_{t-1} - \alpha \frac{\partial L(W)}{\partial W}
\end{equation}
where $\alpha$ is the learning rate equivalent of step size in optimization problems. More details can be found in~\cite{MMCS11} while the background materials presented thus far and those in subsection~\ref{sub:similarity} are the more important aspects with embedded improvements.

\subsection{Instability measure}
\label{sub:similarity}
The similarity index selected for instability measure is the correlation ratio reported in~\cite{AG12} and mathematically proven by~\cite{LCT07} to have low computational requirement as a way of reducing the already-large computation architecture. Also useful was its ability to quantify the relationship between two image frames with differing intensities as well as not directly including the actual images in the computation, hence its choice as our performance metric. Assuming that the input image $X$ has pixels $x \in X$ with intensities $x_i \in (1,255)$ for $i = 1,2,...$ and the inferred output frames denoted as $Y$, the correlation ratio is calculated by computing the total and conditional variances $\sigma$ and $\sigma_i$ in $Y$: %due to the mapping of intensity locations from X to Y %////Please polish this part, this part is a little unclear to me
\begin{equation}
\sigma_i = \frac{1}{Z_i} \sum_{x_i} (Y[x_i])^2 - (\frac{1}{Z_i} \sum Y[x_i])^2
\end{equation}
and:
\begin{equation}
\sigma = \frac{1}{Z} \sum_{X} (Y[x])^2 - (\frac{1}{Z} \sum Y[x])^2
\end{equation}
Using these equations, the correlation ratio is found as:
\begin{equation}
\delta = \frac{1}{Z \sigma} \sum_{i} Z_i \sigma_i
\end{equation}
where $Z_i$  and $Z$ are enumerations of the pixels in $x_i$ and $X$ respectively. Like other chosen measure of dissimilarity such as $\ell_1$ and $\ell_2$ norm, the correlation ratio usually varies from $0$ for uncorrelated images and $1$ for fully correlated images.

%%%%%%%%%%%%%%%%%%%%%%%%%%%%%%%%%%%%%%%%%%%%%%%%%%%%%%%%%%%%%
%%%%%%%%%%%%%%%%%%%%%  Section 4 here %%%%%%%%%%%%%%%%%%%%%%%%%%%%%%%%
\section{Dataset and implementation}
\label{sec:implementation}
\begin{figure*}[!htb]
\centering
  \includegraphics[width=\textwidth,trim={0 0 0 0}]{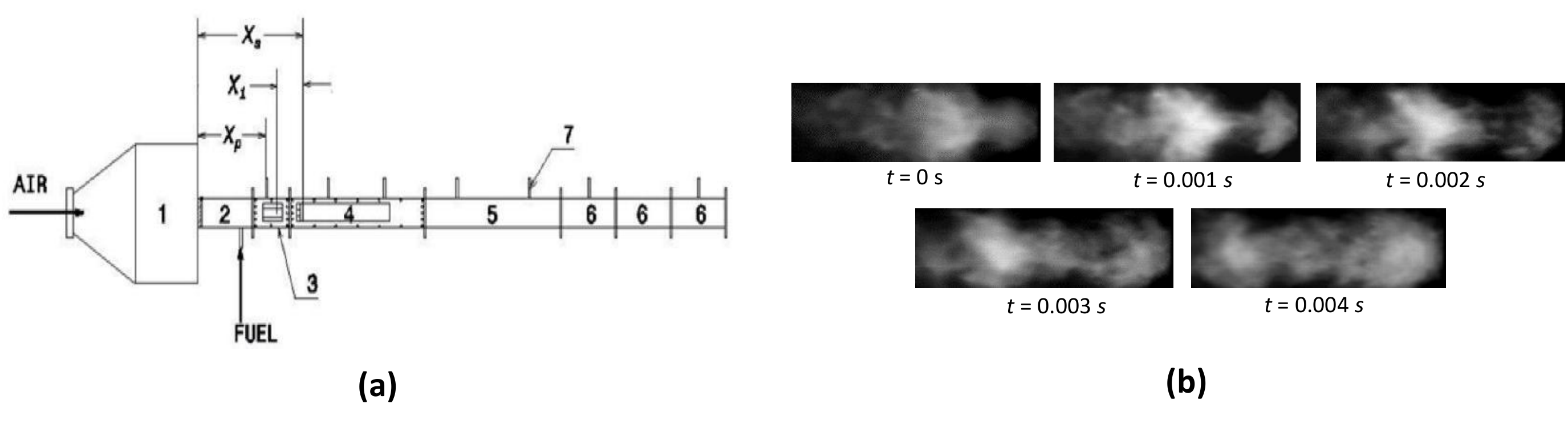}
\caption{a) Schematics of the experimental apparatus. 1 - settling chamber, 2 - inlet duct, 3 - inlet optical access module (IOAM), 4 - test section, 5 $\&$ 6 - big and small extension ducts, 7 - pressure transducers, $X_s$ - swirler location, $X_p$ - transducer port location, $X_i$ - fuel injection location, (b) Visible coherent structure in greyscale images at 900 lpm AFR and full premixing for 45 lpm FFR}
\label{fig:exp_setup}
\end{figure*}

\textbf{Dataset collection and Experimental setup}:
To collect training data for learning coherent structures, thermo-acoustic instability was induced in a laboratory-scale combustor with a 30 mm swirler (60 degree vane angles with geometric swirl number of 1.28). Figure~\ref{fig:exp_setup} (a) shows the setup and a detail description can be found in~\cite{SLSRCPR15}. In the combustor, 4 different instability conditions are induced: 3 seconds of hi-speed videos (i.e., 9000 frames) were captured at 45 lpm (liters per minute) FFR (fuel flow rate) and 900 lpm AFR (air flow rate), and at 28 lpm FFR and 600 lpm AFR for both levels of premixing. Figure~\ref{fig:exp_setup} (b) presents sequences of images of dimension $100\times237$ pixels for unstable ($AFR = 900 lpm$, $FFR = 45 lpm$ and full premixing) state. The flame inlet is on the right side of each image and the flame flows downstream to the left. As the combustion is unstable, figure~\ref{fig:exp_setup} (b) shows formation of mushroom-shaped vortex (coherent structure) at $t = 0, 0.001 s$ and the shedding of that towards downstream from $t = 0.002 s$ to $t = 0.004 s$. For testing the proposed architecture, 5 transition videos of 7 seconds length were collected where stable combustion progressively becomes unstable via 'intermittence' phenomenon (fast switching between stability and instability as a precursor to persistent instability) by reducing FFR or increasing AFR. The transition conditions are as follows (all units are lpm): (i) AFR = 500 and FFR = 40 to 28, (ii)  AFR = 500 and FFR = 40 to 30,  (iii) FFR = 40 and AFR = 500 to 600, (iv) AFR = 600 and FFR = 50 to 35, (v) FFR = 50 and AFR = 700 to 800. For clarity, these data sets are named as $500_{40to38}$, $500_{40to30}$, $40_{500to600}$, $600_{50to35}$, and $50_{700to800}$ respectively for analysis in the subsequent sections of this paper.
\\

\textbf{Training process}:
The architecture inputs and implementation for model learning from data by the architecture are described in this part.
In training the network, $63,000$ gray scale frames having dimensions $100 \times 237$ are resized to $64 \times 64$ for computational simplicity. A total of $35,000$ frames was labeled stable while the remaining $28,000$ were labeled unstable. These images were a combination of datasets with different premixing lengths of either $90$mm or $120$mm and a wide range of air and fuel LPMs for which the combustor is either in a stable or an unstable state. A learning rate of $0.0001$ with momentum = $0.975$ was found to train the model best in the Nesterov based stochastic gradient descent formulation. The network was trained to $100$ epochs in order to conveniently strike a good minima of the validation error. As stated earlier, $\ell^2-$regularization parameters of $0.0001$ each were added to widen the parameter search space for locating the minima by helping to minimize the difference between the test and training. Also, the $\ell^1$ enhances the sparsity of the algorithm while ensuring that only the most likely units are activated.
\begin{figure}[!htb]
\vskip 0.2in
\begin{center}
\centerline{\includegraphics[width=\columnwidth]{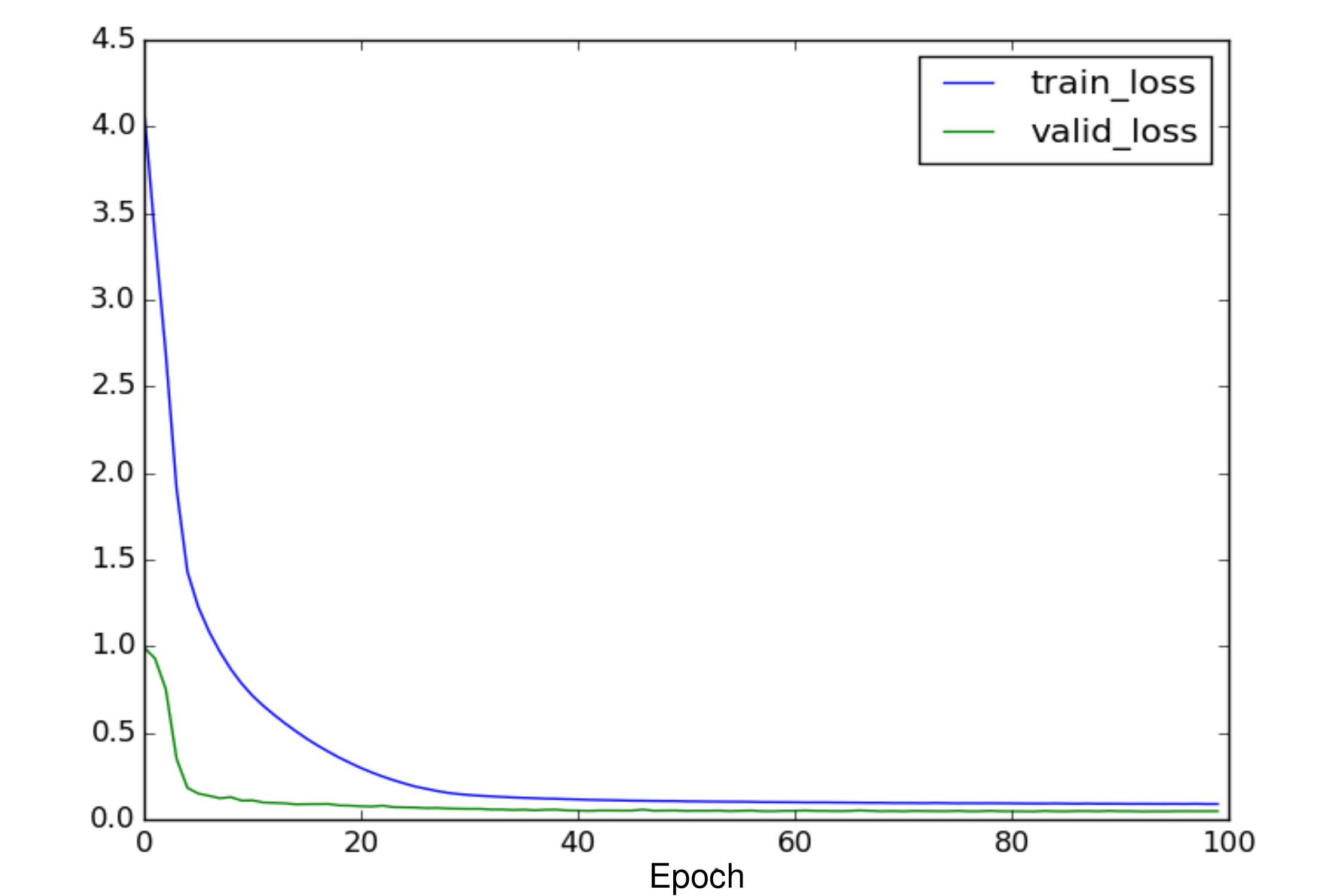}}
\caption{Training and validation losses when training the model.}
\label{fig:loss}
\end{center}
\vskip -0.2in
\end{figure}
Training was done on GPU Titan Black with 2880 CUDA cores, equipped with 6GB video memory, using the python-based machine learning frameworks such as Theano, Lasagne and NoLearn~\cite{BBBLPDTFB10}~\cite{MT15}. Lasagne offers a wide variety of control over the layer types, nonlinearity types, objective functions, interfacing with Theano, and many other features built into it. NoLearn, on the other hand, is a coordinating library for the implementation of the layers in Lasagne which offers model visualization features. While training, a filter of $c \times c$ pixels ($c=3$ in the implementation) and a non-overlapping $p \times p$ ($p$=2) maxpooling were found to be experimentally less costly to produce the results. Algorithm training was done in batches of $128$ training examples which was found to be suitable via cross validation. Note that the batch iterative training in NoLearn and Lasagne functions were replaced with Theano's LeNet 5~\cite{LBBH98} early stopping algorithm which showed further improvements in test performance.
\begin{figure}[!htb]
\vskip 0.2in
\begin{center}
\centerline{\includegraphics[width=0.5\textwidth]{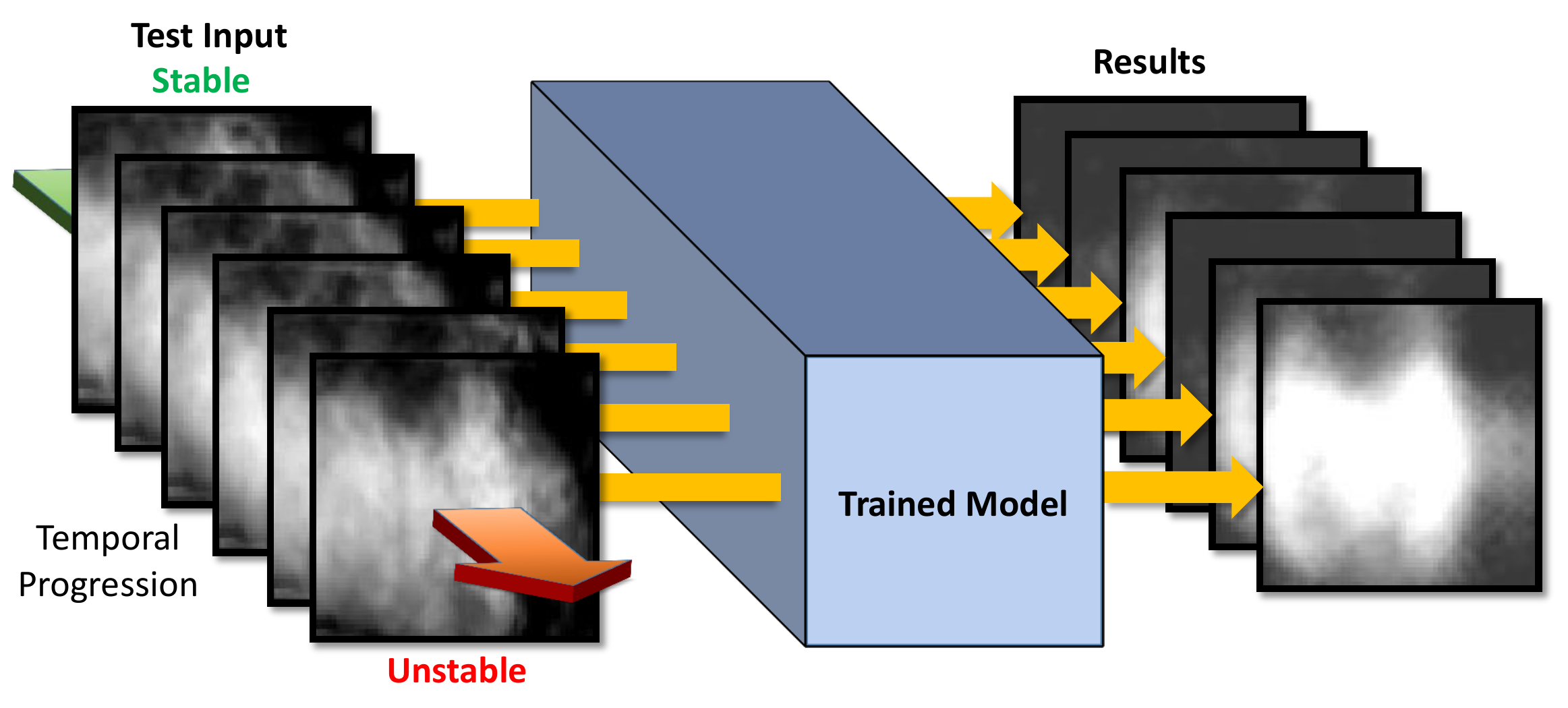}}
\caption{Schematics of implementation of trained network on transition test data}
\label{fig:implement}
\end{center}
\vskip -0.2in
\end{figure}
The architecture in fig.~\ref{fig:cae} shows how the layers are interlinked in the training stage which leads to an overall of $5,090,249$ learnable parameters. The training progress is indicated by the algorithm's loss profile fig.~\ref{fig:loss} showing the validation loss, $valid_loss$, reduction as well as training loss denoted as $train_loss$. Note that the training loss was raised by lowering the regularization parameter in order to allow for more training epochs. This helps to reduce the validation loss more for better result.
%\textbf{Coverage of the training frame at the final layer was estimated at $26.56\%$ and a training capacity of $\approx 70\%$.} \color{blue}(Unclear what is meant by coverage of training frame?)\color{black} .
Leveraging the capability of GPU, generating results from the trained model based on a transition sequence of $21,841$ frames took $ \approx 35.5 secs$.

\section{Results and Discussions}
\label{sec:results}
In this section, some results obtained from the algorithm are discussed and analyzed closely. First, we consider the detection of the presence of region two properties in frames supposedly of region 1 in the early instability detection paradigm. Then we discuss how the network explores the space between the stable and unstable regions to get softer labels assuming the system were static. \\

\begin{figure}[!htb]
\begin{center}
\centerline{\includegraphics[width=0.4\textwidth,trim={0 0 0 0}]{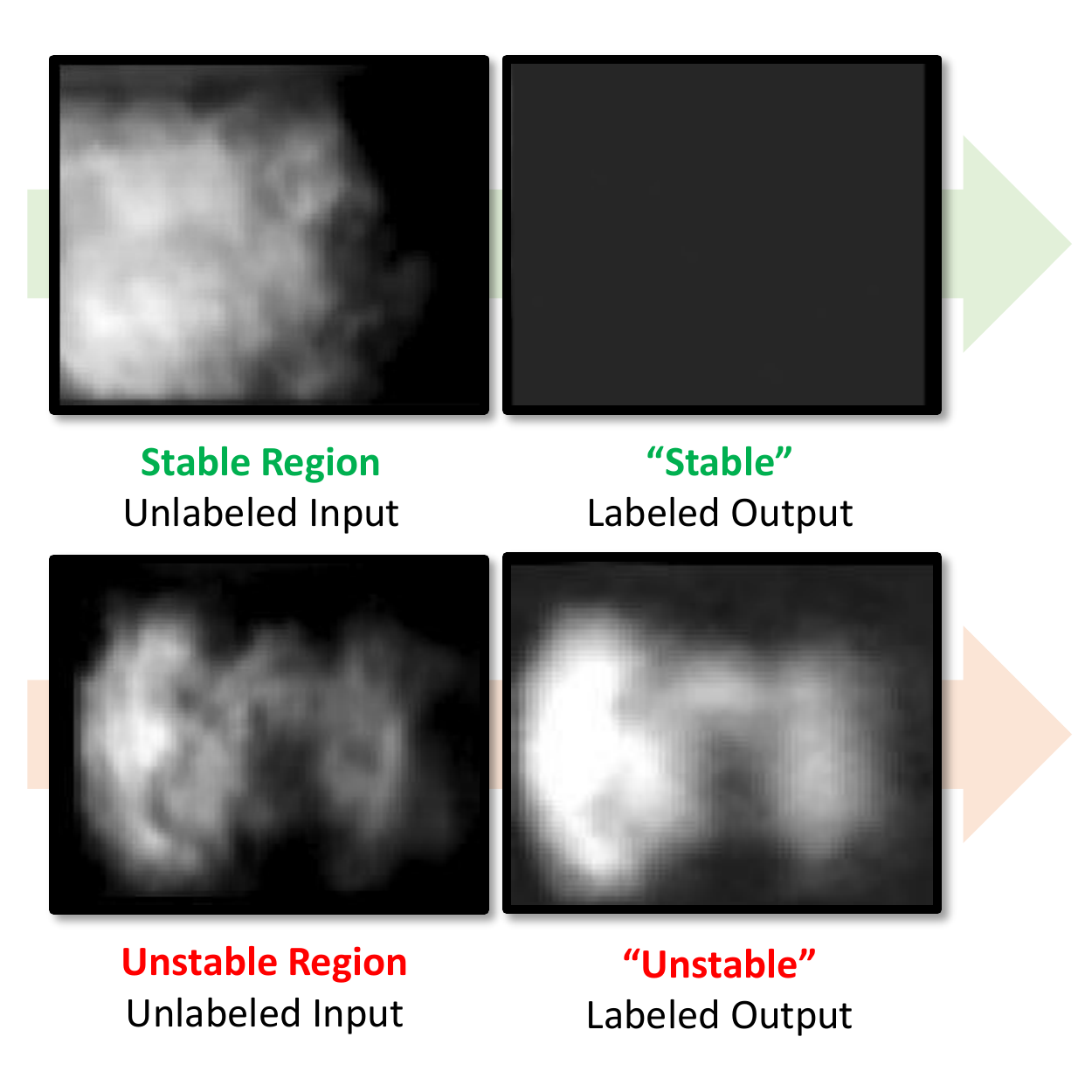}}
\caption{Illustration of the pipeline's ability to reproduce explicit labels.}
\label{fig:explicit}
\end{center}
\vskip -0.1in
\end{figure}

\begin{figure}[!htb]
\begin{center}
\centerline{\includegraphics[width=\columnwidth,trim={0 40 0 40}]{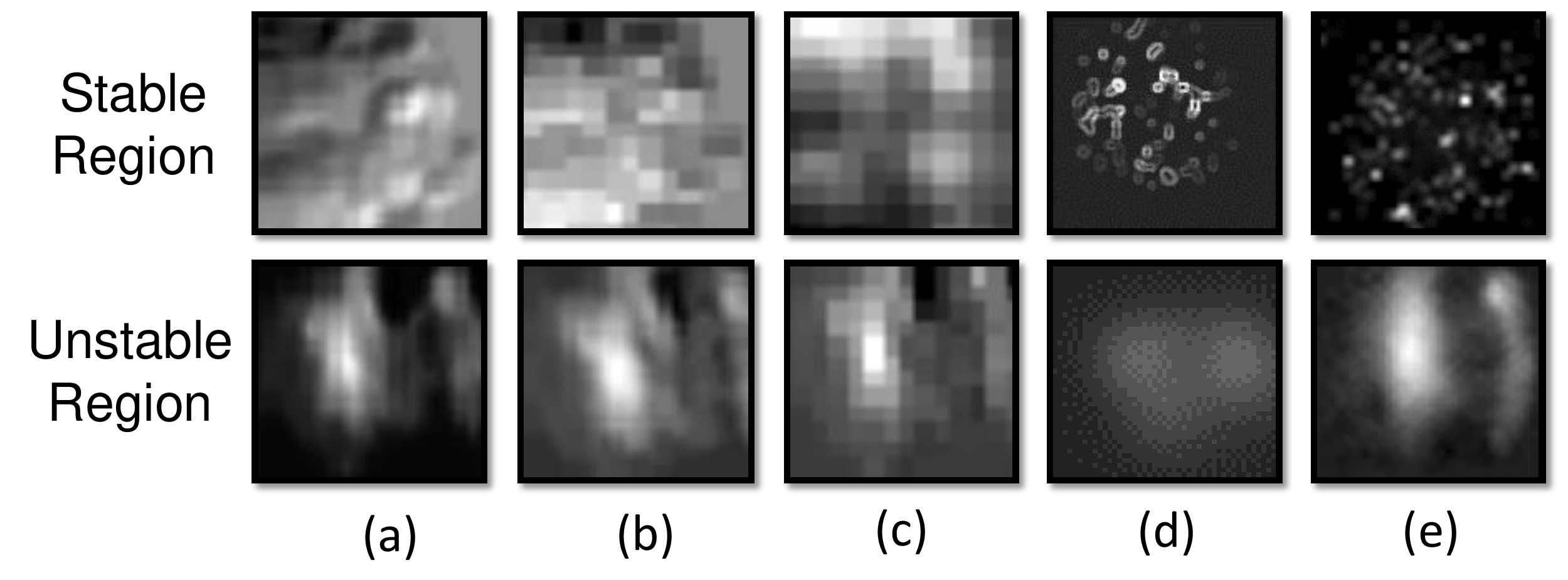}}
\caption{Feature maps for (a) the third convolution layer, (b) the second pooling layer, (c) the fourth convolution layer, (d)  the unpooling layer and (e) the deconvolution layer.}
\label{fig:feature_maps}
\end{center}
\end{figure}

\textbf{Early Detection of combustion instability}:
Let the stable region be denoted by SR on one end of the spectrum and the unstable region be UR on the other end of the spectrum. For emphasis, training of the algorithm was done with explicitly available ground truth labels. These were categorized into frames of stable flame types and frames of unstable flame types. Any unit of frames in the stable region are then given labels of '0' while those of the unstable region were retained in algorithm training. Figure~\ref{fig:explicit} shows the algorithm's ability to reproduce such training in one of the frames trained on.
The algorithm's selective ability is shown by fig.~\ref{fig:explicit}. Feature maps from the model are shown in fig.~\ref{fig:feature_maps} to highlight the detected features and the reconstructed outputs. %With these features, the otherwise black box models becomes gray as the changes and effects towards the final explicit labels of chosen examples are revealed.

Some important feature maps are visualized in in fig.~\ref{fig:feature_maps}. %We note that at the fully connected layers, the weights and biases have the information contents.
The fully connected layers serve at least two important purposes, namely: (1) to reduce further the image dimensions towards only rich explanatory features, and (2) ensuring structural consistency via reshaping to gradually restore the feature maps and output images into dimensions similar to the input.

For frames in the unstable region, the corresponding feature maps showed more activated units responding to the mushroom structures characteristic to unstable combustion. These are highlighted in fig.~\ref{fig:feature_maps}. For those from the stable region, information is seen to be rapidly diffusing from the input into the hidden layers. At each layer, joint parameters capture the trade-off between discarded and retained information from the stable and unstable training sets.
\begin{figure*}[lhtb]
\centering
 \includegraphics[width=1.0\textwidth,trim={0 50 10 0}]{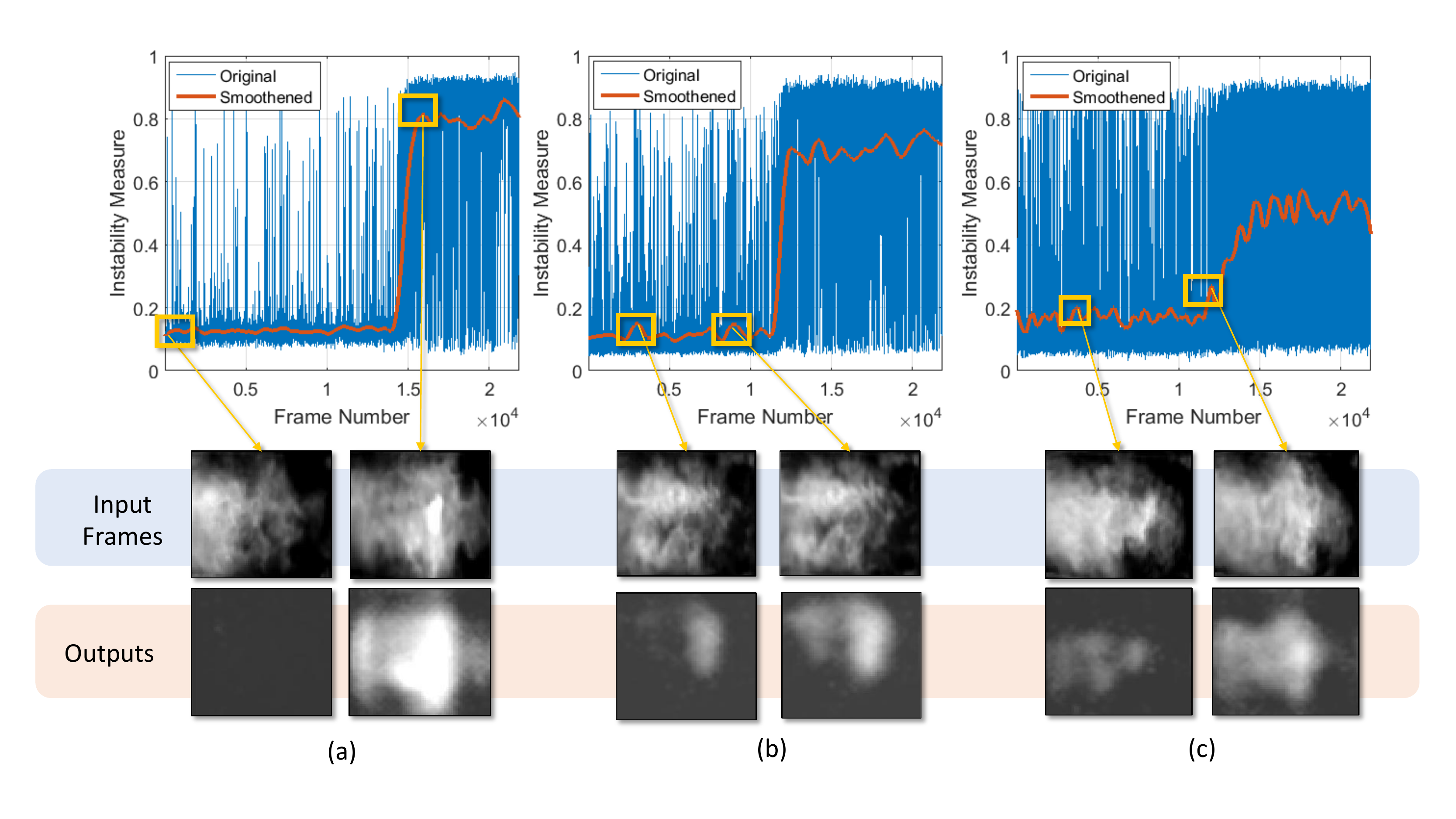}
\caption{Results of transition protocols for: plate a)$600_{50to35}$, b)$500_{40to30}$ and c)$50_{700to800}$ where purple arrows indicate expected results from stability and instability while the green arrows with different intensities indicate the strength of early instability presence in a supposedly stability region}
\label{fig:main_result}
\end{figure*}
Based on the understanding so far, it is hypothesized that the trained model could identify frames having flame types intersecting both stable and unstable regions as discussed in the motivating example. The subsequent parts of the section are devoted to this analysis.\\

After running the test data (the inputs) under different transition conditions through the trained model, the results are presented in in fig.~\ref{fig:main_result}. The figure shows the capability of the model in suppressing unwanted features (the unstable flame images) through masking and revealing the desired anomaly features (the stable flame images).  Note that similarity measure introduced in section~\ref{sub:similarity} was used to evaluate the strength of the algorithm's ability to mask examples closer to the stable region compared to those nearer to the unstable region. Thereafter, a local regression smoothing was applied to obtain weighted moving averages for visualizing the transitional trends.\\

The general trends and fluctuations in instability measures are shown in fig.~\ref{fig:main_result}. These results are similar to those reported in~\cite{NIPS15} where the framework used a neural-symbolic approach with a combination of convolutional neural networks and symbolic time series analysis to obtain instability measures. In fig.~\ref{fig:main_result}(a), output frames from the stable region is suppressed while output frames from the unstable region is entirely visible. Essentially, the model has become a filter which only enables desired features to show up in the outputs. A worthy mention is that intermittency is a precursor to combustion instability. In these regions, the train model produces an output that partially reveals the unstable features as highlighted in fig.~\ref{fig:main_result}(b). The same phenomenon can be observed in fig.~\ref{fig:main_result}(c). As reported in~\cite{NIPS15}, these intermittency phenomena can not be observed from pressure data or POD analysis of image data. Hence the proposed metric localizes the intermittencies prior to full-blown instability with more prominence than other state-of-the-art approaches~\cite{NIPS15}. Better accuracy in tracking intermittence leads to an early detection of instability with less false alarm~\cite{NIPS15}.

\textbf{Frame labeling}:
A computationally complex decision in attempting implicit labeling with these results would be to search through all the frames for adjacent neighbors to a given frame. In clear terms, this means finding which frames come before and after any chosen frame. This kind of search is usually difficult with most primitive low dimensional local labeling algorithms, HMM and MEMM due to dependency depth and labeling bias limitations respectively.\\
\begin{figure*}[lhtb]
\centering
 \includegraphics[width=1.0\textwidth,trim={0 0 0 0}]{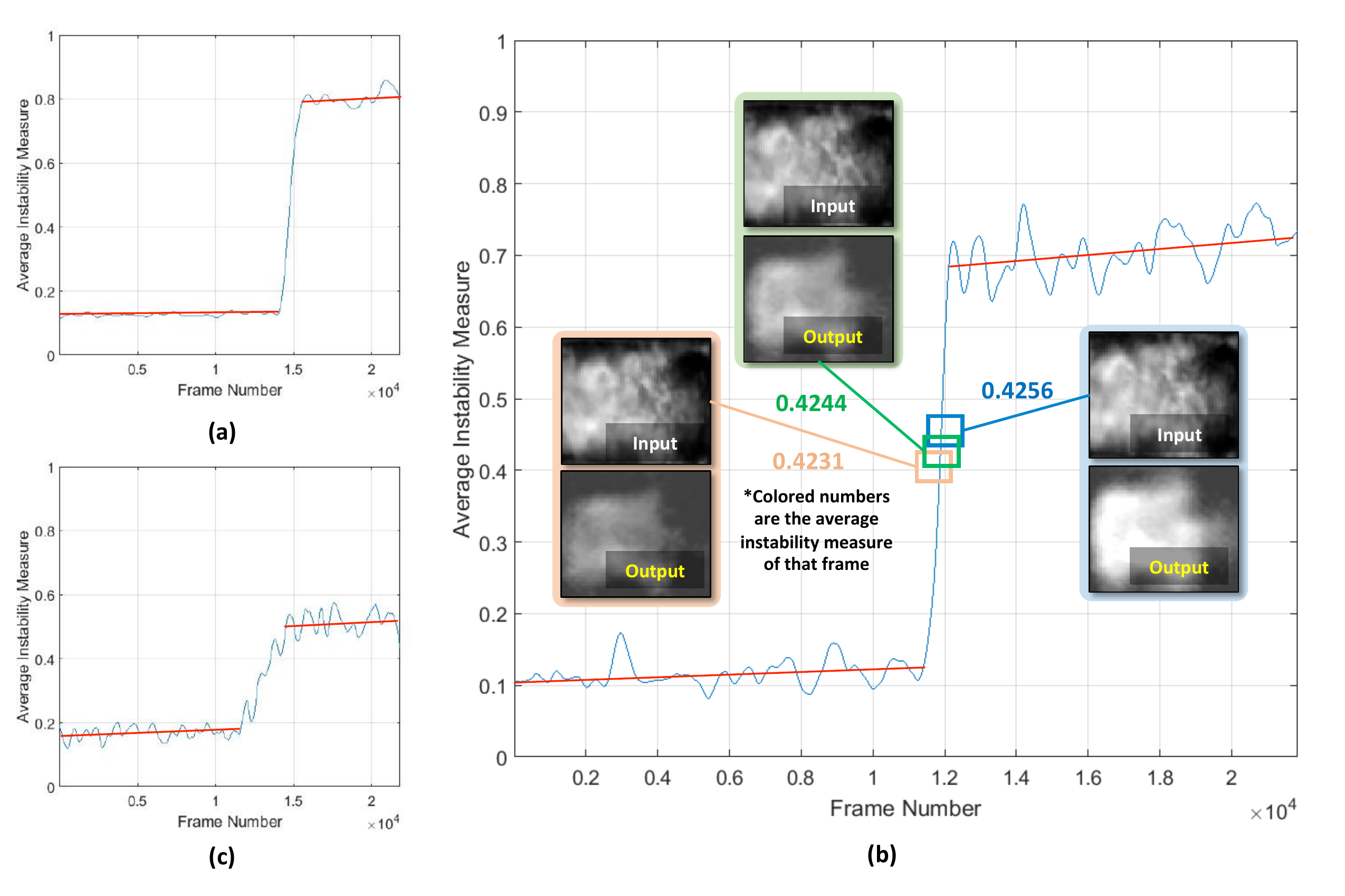}
\caption{Adjacency labeling result of transition protocols for: plate a)$600_{50to35}$, b)$500_{40to30}$ and c) $50_{700to800}$ with three consecutive frames on a transition protocol highlighted}
\label{fig:main_result2}
\end{figure*}
We hypothesize that simplifying any such high dimensional problem using a single value, average instability measure through the composite convolutional auto-encoder would facilitate soft labeling. This is especially required in the regions before and after the abrupt transitioning from SR to UR.  With averagely linear lines superimposed on the frames that are nearer to SR and UR, the complex topology of the frames' arrangement could be simplified to a linear plot that enhances the determination of most likely nearest neighbor frames to any given reference frame.\\
A proposed approach for adjacency labeling in transition protocols of fig.~\ref{fig:main_result} are shown in fig.~\ref{fig:main_result2}. By considering the averaged sections before and after the abrupt transition, the gradual labels could be explored for implicit neighborhood graph decision. Note that the data in this case had to be smoothened to remove the transients introduced by the dynamics of the combustor. However, the richness in the dynamics signifies some defect in this approach for labeling problems of generic applications.\\
Let the red lines be the average linear lines on each examples in fig.~\ref{fig:main_result2} for a static system with similar result to that considered.  The results of transition protocol in fig.~\ref{fig:main_result2}a is closest to the explicit labels example used for training the algorithm because the lines are constant before sudden jump. Also, its neighboring graphs are less graduated than those of fig.~\ref{fig:main_result2}b and fig.~\ref{fig:main_result2}c whose protocols represent more steady rise in the transition property - average instability measure. However, the protocol in fig.~\ref{fig:main_result2}c has more subtlety between the SR and UR with a fuzzy transition region because of the prominence of the second region's property in the first. The closest representation of the implicit labeling problem among the datasets is that shown in fig.~\ref{fig:main_result2}b. Given any frame at random, one would be able to determine all its sets of neighbors in both regions by a graduation of the average instability measure especially after a random shuffling of the frame positions or in cases where the knowledge of ground truth states for each individual frames are not available. Also implicit labeling helps in higher level learning of the flame dynamics from the video by using probabilistic graphical model such as HMM or STSA~\cite{NIPS15}.

Specifically highlighted on plates of fig.~\ref{fig:main_result2}b are the results of three consecutive frames on a region that would represent our hypothesis in the portions before and after full transition for static applications. In this application however, the gradual build up of the two lobes "mushroom-like structure"~\cite{SLSRCPR15} from frame to frame as well as increasing average instability measure are a pointer to the graduated transitioning ability of the framework. Note that the dynamics of the regions in red line would hide this information if frames were selected from there. Thus, we could easily fix frames back to their original position in static applications by the soft labels generated. However, it is expected that the discriminative advantage of the network would intuitively support that presumption. This is a consequence of marginalizing on each hidden layer given the previous layer which has aided determining roughly, neighboring flame patterns and thus providing a coarse to finer labels.

\section{Conclusions and future works}
\label{sec:conclude}
An end-to-end convolutional selective autoencoder is developed to generate fuzzy labels from prior knowledge of hard labeled examples. The framework
is used to perform early detection of combustion instabilities using hi-speed flame video. Validation results are presented using data from a laboratory scale swirl-stabilized combustor. 
%architecture is described and formulated carefully description and was used to confirm results of an interesting combustion problem; that is, to signal in a timely manner the presence of instability in a frames from a swirl-stabilized combustor flame before it eventually occurs. 
The results are discussed in the light of a high fidelity similarity metric which is used to gauge the closeness to ground truth unstable flames. Using the same measure, the architecture was extended to address the neighborhood implicit graph labeling problem. The framework can be generalized for a high-dimensional data in order to perform soft-labeling by interpolation of explicit labels. 
%Despite the potentials of the architecture, a limitation noticed may be its inability to capture cases involving dynamic systems which are usually present in most data.
While the framework is shown to be an efficient diagnostics technique for combustion process in laboratory experiments, large scale validation is underway to demonstrate its wide-range applicability. Apart from that, the framework is also enabling domain experts to learn more about the coherent structures that appear during combustion instabilities. From a technical point of view, future research will involve extending the framework to multi-class implicit labeling problems. 

\section{Acknowledgments}
Authors sincerely acknowledge the extensive data collection performed by Vikram Ramanan and Dr. Satyanarayanan Chakravarthy at Indian Institute of Technology Madras (IITM), Chennai. Authors also gratefully acknowledge the support of NVIDIA Corporation with the donation of the GeForce GTX TITAN Black GPU used for this research.

\bibliographystyle{abbrv}

\bibliography{sigproc}  % sigproc.bib is the name of the Bibliography in this case

\begin{thebibliography}{10}

\bibitem{AEB06}
M.~Aharon, M.~Elad, and A.~Bruckstein.
\newblock An algorithm for designing overcomplete dictionaries for sparse
  representation.
\newblock {\em IEEE Transactions on Signal Processing}, 54(11):4311--4322,
  2006.

\bibitem{YB08}
Y.~Bengio.
\newblock Learning deep architecture for ai.
\newblock {\em Foundations and Trends in Machine Learning}, pages 1--71, 2008.

\bibitem{BPP96}
A.~L. Berger, S.~A.~D. Pietra, and V.~J.~D. Pietra.
\newblock A maximum entropy approach to natural language processing.
\newblock {\em Association for Computational Linguistics}, 22(1):1 -- 36, 1996.

\bibitem{BBBLPDTFB10}
J.~Bergstra, O.~Breulex, F.~Bastien, P.~Lamblin, R.~Pascanu, G.~Desjardins,
  J.~Turian, D.~Warde-Farley, and Y.~Bengio.
\newblock Theano: a cpu and gpu math expression compiler.
\newblock {\em Proceedings of the Python for Scientific Computing Conference
  (SciPy)}, June 2010.
\newblock Oral Presentation.

\bibitem{BHL93}
G.~Berkooz, P.~Holmes, and J.~L. Lumley.
\newblock The proper orthogonal decomposition in the analysis of turbulent
  flows.
\newblock {\em Annual Review of Fluid Mechanics}, 25(1):539--575, 1993.

\bibitem{B06}
C.~M. Bishop.
\newblock {\em Pattern Recognition and Machine Learning}.
\newblock Springer, New York, NY, USA, 2006.

\bibitem{BC12}
B.~Cheung.
\newblock Convolutional neural networks applied to human face classification.
\newblock {\em ICMLA}, 2(12):580--583, 2012.

\bibitem{CW08}
R.~Collobert and J.~Weston.
\newblock A unified architecture for natural language processing:deep neural
  networks with multitask learning.
\newblock {\em 25th International Conference on Machine Learning}, pages 1 --
  7, 2008.

\bibitem{DHS11}
J.~Duchi, E.~Hazan, and Y.~Singer.
\newblock Adaptive subgradient methods for online learning and stochastic
  optimization.
\newblock {\em JMLR}, 12:2121--2159, July 2011.

\bibitem{HE10}
H.~Erdogan.
\newblock A tutorial on sequence labeling.
\newblock ICMLA, December 2010.

\bibitem{FCNL13}
C.~Farabet, C.~Couprie, L.~Najman, and Y.~LeCun.
\newblock Learning hierarchical features for scene labeling.
\newblock {\em exdb}, pages 1--15, 2013.

\bibitem{FM04}
J.~Fung and S.~Mann.
\newblock Using multiple graphics cards as a general purpose parallel computer:
  Applications to compute vision.
\newblock 1:805--808, August 2004.

\bibitem{AG12}
A.~A. Goshtasby.
\newblock {\em Similarity and Dissimilarity Measures}, chapter~2, pages 1 --
  66.
\newblock Number 978-1-4471-2457-3. Springer-Verlag London Limited, 2012.

\bibitem{AG14}
A.~Graves.
\newblock Generating sequences with recurrent neural networks.
\newblock {\em arXiv:1308.0850v5 [cs.NE]}, pages 1 -- 43, June 2014.

\bibitem{AGJS05}
A.~Graves and J.~Schmidhuber.
\newblock Framewise phoneme classification with bidirectional lstm and other
  neural network architectures.
\newblock {\em IJCNN}, pages 1 -- 8, 2005.

\bibitem{H83}
A.~K. M.~F. Hussain.
\newblock Coherent structures - reality and myth.
\newblock {\em Physics of Fluids}, 26(10):2816--2850, 1983.

\bibitem{SJ15}
S.~Jones.
\newblock Convolutional autoencoders in python/theano/lasagne, April 2015.

\bibitem{KSH12}
A.~Krizhevsky, I.~Sutskever, and G.~E. Hinton.
\newblock Imagenet classification with deep convolutional neural networks.
\newblock In {\em NIPS}, 2012.

\bibitem{KACFC14}
T.~Kulesza, S.~Amershi, R.~Caruana, D.~Fisher, and D.~Charles.
\newblock Strudtured labeling to facilitate concept evolution in machine
  learning.
\newblock {\em ACM}, pages 1 -- 10, April 2014.

\bibitem{LMP01}
J.~Lafferty, A.~McCallum, and F.~C. Pereira.
\newblock Conditional random fields: Probabilistic models for segmenting and
  labeling sequence data.
\newblock {\em International Conference on Machine Learning}, pages 282--289,
  June 2001.

\bibitem{LB95}
Y.~LeCun and Y.~Bengio.
\newblock {\em Convolutional networks for Images, Speech and Time-Series},
  1998.

\bibitem{LBBH98}
Y.~LeCun, L.~Bottou, Y.~Bengio, and P.~Haffner.
\newblock Gradient-based learning applied to document recognition.
\newblock {\em Proc of IEEE}, pages 1--46, November 1998.

\bibitem{LCT07}
D.~Lewandowski, R.~M. Cooke, and R.~J.~D. Tebbens.
\newblock Sample-based estimation of correlation ratio with polynomial
  approximation.
\newblock {\em ACM}, V(N):1--16, May 2007.

\bibitem{MMCS11}
J.~Masci, U.~Meier, D.~Ciresan, and J.~Schmidhuber.
\newblock {\em Stacked Convolutional Auto-Encoders for Hierarchical Feature
  Extraction}, pages 52--59.
\newblock Number 6791. Springer-Verlag Berlin Heidelberg, 2011.

\bibitem{AM12}
A.~Meyer.
\newblock Hmm and part of speech tagging.
\newblock Lecture Note, 2011-2012.

\bibitem{R89}
L.~Rabiner.
\newblock A tutorial on hidden markov models and selected applications in
  speech proccessing.
\newblock {\em Proceedings of the {IEEE}}, 77(2):257--286, 1989.

\bibitem{NIPS15}
S.~Sarkar, K.~G. Lore, and S.~Sarkar.
\newblock Early detection of combustion instability by neural-symbolic analysis
  on hi-speed video.
\newblock In {\em Workshop on Cognitive Computation: Integrating Neural and
  Symbolic Approaches (CoCo @ NIPS 2015)}, Montreal, Canada, December 2015.

\bibitem{SLSRCPR15}
S.~Sarkar, K.~G. Lore, S.~Sarkar, V.~Ramaman, S.~R. Chakravarthy, S.~Phoha, and
  A.~Ray.
\newblock Early detection of combustion instability from hi-speed flame images
  via deep learning and symbolic time series analysis.
\newblock {\em Annual Conference of the Prognostics and Health Management
  Management Society}, pages 1--10, 2015.

\bibitem{SMB10}
D.~Scherer, A.~Muller, and S.~Behnke.
\newblock Evaluation of pooling operations in convolutional architectures for
  object recognition.
\newblock {\em Intenational Conference on Artificial Neural Networks, ICANN},
  pages 1--10, 2010.

\bibitem{S10}
P.~J. Schmid.
\newblock Dynamic mode decomposition of numerical and experimental data.
\newblock {\em Journal of Fluid Mechanics}, 656:5--28, 2010.

\bibitem{SPKL15}
T.~Sercu, C.~Puhrsch, B.~Kingsbury, and Y.~LeCun.
\newblock Very deep multilingual convolutional neural networks for lvcsr.
\newblock {\em arXiv:1509.08967v1 [cs.CL]}, page~5, September 2015.

\bibitem{SMDH13}
I.~Sutskever, J.~Martens, G.~Dahl, and G.~Hinton.
\newblock On the importance of initialization and momentum in deep learning.
\newblock {\em International Conference on Machine Learning, JMLR}, 28, 2013.

\bibitem{TKGG11}
J.~B. Tenenbaum, C.~Kemp, T.~L. Griffiths, and N.~D. Goodman.
\newblock How to grow a mind: Statistics, structure, and abstraction.
\newblock {\em Science}, 331:1279--1285, 2011.

\bibitem{MT15}
M.~Thoma.
\newblock Lasagne for python newbies, February 2016.

\bibitem{MZ12}
M.~D. Zeiler.
\newblock Adadelta:an adaptive learning rate method.
\newblock {\em arXiv:1212.5701v1}, pages 1--6, December 2012.

\end{thebibliography}
%\balancecolumns % GM June 2007
% That's all folks!
\end{document}